\documentclass[twoside,11pt]{article}

\usepackage{jmlr2e}

\usepackage{booktabs}
\usepackage{multirow}
\usepackage{tabularx}
\usepackage{amsmath}

\usepackage{listings}
\usepackage{pythonhighlight}
\usepackage{wrapfig}

\usepackage{lastpage}
\jmlrheading{21}{2020}{1-\pageref{LastPage}}{5/19; Revised
1/20}{2/20}{19-429}{Jian Guo, He He, Tong He, Leonard Lausen, Mu Li, Haibin Lin, Xingjian Shi, Chenguang Wang, Junyuan Xie, Sheng Zha, Aston Zhang, Hang Zhang, Zhi Zhang, Zhongyue Zhang, Shuai Zheng, Yi Zhu}
\ShortHeadings{GluonCV and GluonNLP: Deep Learning in CV and NLP}{Guo, He, He, Lausen, Li, Lin, Shi, Wang, Xie, Zha, Zhang $\times 4$, Zheng, Zhu}

\firstpageno{1}

\begin{document}

\title{GluonCV and GluonNLP: Deep Learning in\\Computer Vision and Natural Language Processing}

\author{\name Jian Guo  \email gjian@umich.edu\\
			\addr University of Michigan MI, USA
			\AND
			\name He He  \email hehea@amazon.com\\
			\name Tong He \email htong@amazon.com\\
       		\name Leonard Lausen \email lausen@amazon.com \\
       	\thanks{Mu Li is the corresponding author.}
			\name Mu Li \email mli@amazon.com \\
			\name Haibin Lin \email haibilin@amazon.com\\
			\name Xingjian Shi \email xjshi@amazon.com \\
			\name Chenguang Wang \email chgwang@amazon.com\\
			\name Junyuan Xie \email eric.jy.xie@gmail.com\\
			\name Sheng Zha  \email zhasheng@amazon.com\\
			\name Aston Zhang  \email astonz@amazon.com\\
			\name Hang Zhang  \email hzaws@amazon.com\\
			\name Zhi Zhang  \email zhiz@amazon.com\\
			\name Zhongyue Zhang  \email zhongyue@amazon.com\\
			\name Shuai Zheng \email shzheng@amazon.com \\
			\name Yi Zhu  \email yzaws@amazon.com\\
			\addr Amazon Web Services, CA, USA}
\editor{Antti Honkela}

\maketitle

\begin{abstract}
We present GluonCV and GluonNLP, the deep learning toolkits for computer vision and natural language processing based on Apache MXNet (incubating). These toolkits provide state-of-the-art pre-trained  models,  training scripts, and training logs, to facilitate rapid prototyping and promote reproducible research. We also provide modular APIs with flexible building blocks to enable efficient customization. Leveraging the MXNet ecosystem, the deep learning models in GluonCV and GluonNLP can be deployed onto a variety of platforms with different programming languages. The Apache 2.0 license has been adopted by GluonCV and GluonNLP to allow for software distribution, modification, and usage.
\end{abstract}

\begin{keywords}
Machine Learning, Deep Learning, Apache MXNet, Computer Vision, Natural Language Processing
\end{keywords}

\section{Introduction}

Deep learning, a sub-field of machine learning research, has driven the rapid progress in artificial intelligence research, leading to astonishing breakthroughs on long-standing problems in a plethora of fields such as computer vision and natural language processing. Tools powered by deep learning are changing the way movies are made~\citep{knorr2018deepstereobrush}, diseases are diagnosed~\citep{lipton2015learning}, and play a growing role in understanding and communicating with humans~\citep{zhu2018sdnet}.

Such development is made possible by deep learning frameworks, such as Caffe~\citep{jia2014caffe}, Chainer~\citep{tokui2015chainer}, CNTK~\citep{seide2016cntk}, Apache (incubating) MXNet~\citep{chen2015mxnet}, PyTorch~\citep{paszke2017automatic}, TensorFlow~\citep{abadi2016tensorflow}, and Theano~\citep{bastien2012theano}. These frameworks have been crucial in disseminating ideas in the field. Specifically, imperative tools, arguably spearheaded by Chainer, are easy to learn, read, and debug. 
Such benefits make imperative programming interface quickly adopted by the Gluon API of MXNet (though it can be seamlessly switched to symbolic programming for high performance), PyTorch, and TensorFlow Eager. 

Leveraging the imperative Gluon API of MXNet, we design and develop the GluonCV and GluonNLP (referred to as GluonCV/NLP hereinafter) toolkits for deep learning in computer vision and natural language processing. GluonCV/NLP simultaneously
i)  provide modular APIs to allow customization by re-using efficient building blocks;
ii) provide pre-trained state-of-the-art models, training scripts, and training logs to enable fast prototyping and promote reproducible research;
iii) provide models that can be deployed in a wide variety of programming languages including C++, Clojure, Java, Julia, Perl, Python, R, and Scala (via the MXNet ecosystem).

Sharing the same MXNet backend and enjoying its benefits such as multiple language bindings, which are not offered by other frameworks, the frontend of GluonCV/NLP are separated to allow users of either computer vision or natural langauge processing to download and install either GluonCV or GluonNLP of a smaller size than that of their combination.

\section{Design and Features}

In the following, we describe the design and features of GluonCV/NLP.

\subsection{Modular APIs}


GluonCV/NLP provide access to modular APIs to allow users to customize their model design, training, and inference by re-using efficient components across different models.
Such common components include (but are not limited to) data processing utilities, models with individual components, initialization methods, and loss functions.

To elucidate how the modular API facilitates efficient implementation, let us take the \texttt{data} API of GluonCV/NLP as an example, which is used to build efficient data pipelines with popular benchmark data sets or those supplied by users. 
In computer vision and natural language processing tasks, inputs or labels often come in with different shapes, such as images with a varying number of objects and sentences of different lengths.
Thus,
the \texttt{data} API provides a collection of utilities to sample inputs or labels then transform them into mini-batches to be efficiently computed. 

The following code snippet shows an example of using fixed bucketing to sample mini-batches more efficiently. In the first assignment statement, the \texttt{batchify} utilities provided by the \texttt{data} API specifies that inputs of different shapes will be padded to have the same length then be stacked as a mini-batch. The following \texttt{FixedBucketSampler} class of the \texttt{data} API will group inputs of \emph{more similar} shapes into the same mini-batch so more computation is saved due to less padding.
In the end, we pass the aforementioned specifications, together with the actual data set \texttt{train\_data}, via arguments of the \texttt{DataLoader} class of the Gluon API of MXNet, so it will return efficient mini-batches of \texttt{train\_data}.

\begin{python}
import gluon, gluonnlp
batchify_fn = gluonnlp.data.batchify.Tuple(
    gluonnlp.data.batchify.Pad(), gluonnlp.data.batchify.Stack())
train_sampler = gluonnlp.data.FixedBucketSampler(
    lengths=train_data.transform(lambda x: len(x[0])),
    batch_size=batch_size, shuffle=True)
train_iter = gluon.data.DataLoader(
    train_data, batchify_fn=batchify_fn, batch_sampler=train_sampler)
\end{python}

Besides, users can access a wide range of popular data sets via the \texttt{data} API, including (but are not limited to) ImageNet of image classification, VOC of object detection, COCO of instance segmentation, SST of sentiment analysis,
IWSLT of machine translation,
SQuAD of question answering, and
WikiText of language modeling.
The code snippet below shows that users can access training sets of IWSLT2015 English-Vietnamese and SQuAD 2.0, and the test set of WikiText103 with just one line of code via the \texttt{data} API.

\begin{python}
import gluonnlp
iwslt15 = gluonnlp.data.IWSLT2015('train', src_lang='en', tgt_lang='vi')
squad = gluonnlp.data.SQuAD('train', '2.0')
wikitext103 = gluonnlp.data.WikiText103('test')
\end{python}

\begin{wrapfigure}{r}{0.45\textwidth}
    \centering
    \includegraphics[width=0.45\textwidth]{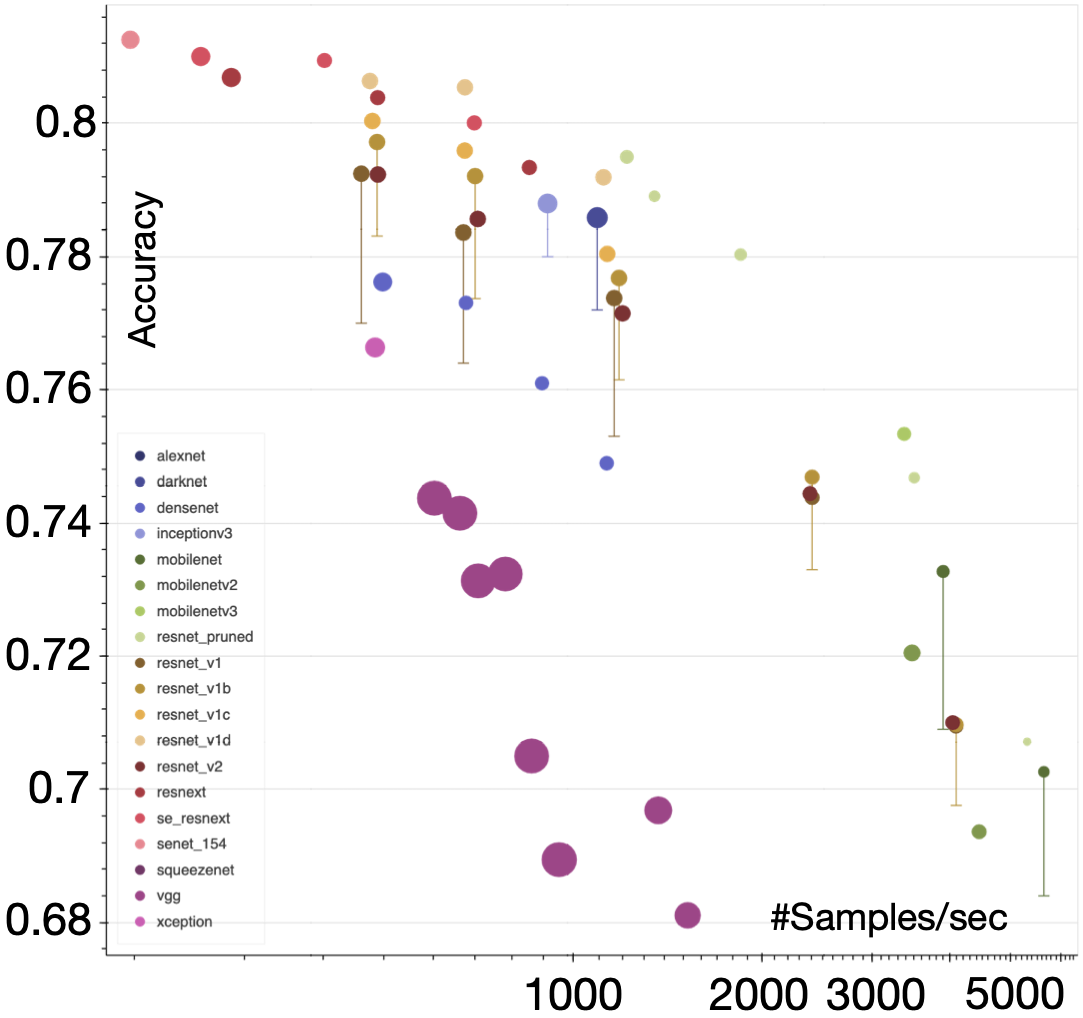}
    \caption{GluonCV's inference throughputs vs. validation accuracy. Circle Area is proportional to device memory requirement.}
    \vspace{-30pt}
    \label{fig:acc_throughput}
\end{wrapfigure}

\subsection{Model Zoo}

Building upon those modular APIs, GluonCV/NLP provide pre-trained state-of-the-art models, training scripts, and training logs via the model zoo to enable fast prototyping and promote reproducible research.
As of the time of writing, GluonCV/NLP have provided over 200 models for common computer vision and natural language processing tasks, such as image classification, object detection, semantic segmentation, instance segmentation, pose estimation, video action recognition, word embedding, language model, machine translation, sentiment analysis, natural language inference, dependency parsing, and question answering.
Figure~\ref{fig:acc_throughput} visualizes inference throughputs (on 1 NVIDIA TESLA V100) vs. validation accuracy of ImageNet pre-trained models on image classification.

\subsection{Leveraging the MXNet Ecosystem}

GluonCV/NLP have benefitted from the MXNet ecosystem through use of MXNet.
At the lowest level, MXNet provides high-performance C++ implementations of operators that are leveraged by GluonCV/NLP; thus, improvements in low-level components of MXNet often result in performance gains in GluonCV/NLP.
Same as any other model implemented with MXNet, GluonCV/NLP can be used to train models on CPU, GPU (single or multiple), and multiple machines.
In sharp contrast to building upon other deep learning frameworks,
through the unique hybridizing mechanism by MXNet~\citep{zhang-et-al-2019},
usually GluonCV/NLP models can be deployed with no or minimal configuration in a wide spectrum of programming languages including C++, Clojure, Java, Julia, Perl, Python, R, and Scala.
There are also ongoing efforts to bring more quantization (int8 and float16 inference) benefits from MXNet to GluonCV/NLP to further accelerate model inference,
e.g., comparing with the float32 inference,
our sentence classification result on the MRPC dataset shows that the int8 inference reduces latency of the $\text{BERT}_{\text{BASE}}$  model by 59.6\% on Intel CLX-8280.

The documentation \url{https://gluon-cv.mxnet.io/} and \url{http://gluon-nlp.mxnet.io/} of GluonCV/NLP include installation instructions, contribution instructions, open source repositories, extensive API reference, and comprehensive tutorials.
As another benefit of leveraging the MXNet ecosystem, the GluonCV/NLP documentation is supplemented by the interactive open source book \emph{Dive into Deep Learning}~\citep{zhang-et-al-2019}, which provides sufficient background knowledge about GluonCV/NLP tasks, models, and building blocks. Notably, some users of \emph{Dive into Deep Learning} have later become contributors of GluonCV/NLP.

\subsection{Requirement, Availability, and Community}

GluonCV/NLP are implemented in Python and are available for systems running Linux, macOS, and Windows since Python is platform agnostic.
The minimum and open source package (\emph{e.g.}, MXNet) requirements are specified in the documentation.
As of the time of writing, GluonCV/NLP have reached version 0.6 and 0.8 respectively, and have been open sourced under the Apache 2.0 license.
The rapid growth in the vibrant open source community has spurred active development of new features in the toolkits. 
Each pull request (code contribution) will trigger the continuous integration server to run all the test cases in the code repositories.
All the changes based on community feedback can be found through the merged pull requests.
For example, the open-source community has enriched the \texttt{data} API by contributing more utilities, and added implementations, training scripts, or training logs of a variety of models, such as textCNN and CycleGAN.

\section{Performance}

We demonstrate the performance of GluonCV/NLP models in various computer vision and natural language processing tasks.
Specifically, we evaluate popular or state-of-the-art models on standard benchmark data sets.
In the experiments, we compare model performance between GluonCV/NLP and other open source implementations with Caffe, Caffe2, Theano, and TensorFlow, including ResNet~\citep{he2016deep} and MobileNet~\citep{howard2017mobilenets} for image classification (ImageNet), Faster R-CNN~\citep{girshick2015fast} for object detection (COCO), Mask R-CNN~\citep{he2017mask} for instance segmentation, Simple Pose~\citep{xiao2018simple} for pose estimation (COCO), Inflated 3D networks (I3D)~\citep{carreira2017i3d} for video action recognition, textCNN~\citep{kim2014convolutional} for sentiment analysis (TREC), and BERT~\citep{devlin2018bert} for question answering (SQuAD 1.1), sentiment analysis (SST-2), natural langauge inference (MNLI-m), and paraphrasing (MRPC). Table~\ref{table:perf} shows that the GluonCV/GluonNLP implementation  matches or outperforms the compared open source implementation for the same model evaluated on the same data set.
In some cases, such as image classification with ResNet-50 on the ImageNet data set,
our implementation has significantly outperformed implementations in other frameworks.
We highlight that the outperformance can be attributed to 
``minor" refinements such as in data augmentations and optimization methods in the training procedure~\citep{he2019bag}.

\begin{table*}[t!]
\caption{Comparison of model performance (in percentage) on the validation data sets between GluonCV/NLP and other open source implementations (OOSI) across popular computer vision and natural language processing tasks and data sets.}
\label{table:perf}
\centering
\scriptsize
\begin{tabular}{llll|cc}
\toprule
Task & Data set & Model & Measure & GluonCV/NLP & OOSI\\
\midrule
\textbf{Image Classification}    &ImageNet	& ResNet-50    &top-1 acc. &79.2&$75.3^{[\text{a}]}$\\
\textbf{Image Classification}     &ImageNet		&ResNet-101    &top-1 acc.        &80.5&$76.4^{[\text{a}]}$\\
\textbf{Image Classification}        &ImageNet	& MobileNet 1.0	         &top-1 acc.             &73.3&$70.9^{[\text{b}]}$\\
\textbf{Object Detection}        &COCO	& Faster R-CNN	         &mAP             &40.1&$39.6^{[\text{c}]}$\\
\textbf{Instance Segmentation}        &COCO	& Mask R-CNN	         &mask AP             &33.1&$32.8^{[\text{c}]}$\\
\textbf{Pose Estimation}        &COCO	& Simple Pose (f)	         &OKS AP            &74.2&N.A.\\
\textbf{Action Recognition}        &Kinetics400	& I3D ResNet-50      & top-1 acc.            &74.0& $72.9^{[\text{f}]}$ \\
\textbf{Sentiment Analysis}        &TREC	& textCNN         &acc.             &92.8& $92.2^{[\text{e}]}$\\
\textbf{Sentiment Analysis}        &SST-2	& $\text{BERT}_{\text{BASE}}$         &acc.             &93.0& $92.7^{[\text{e}]}$\\
\textbf{Question Answering}        &SQuAD 1.1	& $\text{BERT}_{\text{BASE}}$         &F1/EM             &88.5/81.0&88.5/$80.8^{[\text{e}]}$\\
\textbf{Question Answering}        &SQuAD 1.1	& $\text{BERT}_{\text{LARGE}}$         &F1/EM             &91.0/84.1&90.9/$84.1^{[\text{e}]}$\\
\textbf{Natural Language Inference}        &MNLI-m	& $\text{BERT}_{\text{BASE}}$         &acc.             &84.6& $84.4^{[\text{e}]}$\\
\textbf{Paraphrasing}        &MRPC	& $\text{BERT}_{\text{BASE}}$         &acc.             &88.7& $86.7^{[\text{e}]}$\\
\bottomrule
\multicolumn{6}{l}{[a] \url{https://github.com/KaimingHe/deep-residual-networks} (in Caffe)}\\
\multicolumn{6}{l}{[b] \url{https://github.com/tensorflow/models/blob/master/research/slim/nets/mobilenet_v1.md} (in TensorFlow)}\\
\multicolumn{6}{l}{[c] \url{https://github.com/facebookresearch/Detectron} (in Caffe2)}\\
\multicolumn{6}{l}{[d] \url{https://github.com/yoonkim/CNN_sentence} (in Theano)}\\
\multicolumn{6}{l}{[e] \url{https://github.com/google-research/bert} (in TensorFlow)}\\
\multicolumn{6}{l}{[f] \url{https://github.com/open-mmlab/mmaction} (in PyTorch)}\\
\end{tabular}
\end{table*}

\section{Conclusion}

GluonCV/NLP provide modular APIs and the model zoo to allow users to rapidly try out new ideas or develop downstream applications in computer vision and natural language processing. GluonCV/NLP are in active development and our future works include further enriching the API and the model zoo, 
accelerating model inference,
improving compatibility with the NumPy interface, 
and supporting deployment in more scenarios.


\acks{We would like to thank all the contributors of GluonCV and GluonNLP (the \texttt{git log} command can be used to list all the contributors).
Specifically, we thank Xiaoting He, Heewon Jeon, Kangjian Wu, and Luyu Xia for providing part of results in Table~\ref{table:perf}. We would also like to thank the entire MXNet community for their foundational contributions.}


\newpage

\vskip 0.2in
\bibliography{ref}

\end{document}